\documentclass[12pt]{article}
\usepackage{scicite}

\usepackage[letterpaper,margin=1in]{geometry}

\renewenvironment{abstract}
	{\quotation}
	{\endquotation}

\date{}

\makeatletter
\renewcommand{\fnum@figure}{\textbf{Figure \thefigure}}
\renewcommand{\fnum@table}{\textbf{Table \thetable}}
\makeatother

\usepackage{graphicx}%
\usepackage{multirow}%
\usepackage{amsmath,amssymb,amsfonts}%
\usepackage{amsthm}%
\usepackage{hyperref}
\usepackage{cleveref}
\usepackage{mathrsfs}%
\usepackage[title]{appendix}%
\usepackage{textcomp}%
\usepackage{manyfoot}%
\usepackage{booktabs}%
\usepackage{algorithm}%
\usepackage{algorithmicx}%
\usepackage{algpseudocode}%
\usepackage{listings}%
\usepackage{gensymb}
\usepackage[english]{babel}
\usepackage[normalem]{ulem}

\addto\extrasenglish{
  
}
\addto\extrasenglish{
  
}
\addto\extrasenglish{
  
}
\addto\extrasenglish{
  
}
\addto\extrasenglish{
  
}
\addto\extrasenglish{
  
}
\addto\extrasenglish{
  
}
\usepackage[table]{xcolor}
\definecolor{MColor}{rgb}{0.2. 0.8, 0.2}

\newcommand{\facebot}{\textit{face-bot} }
\newcommand{\gridbot}{\textit{grid-bot} }
\newcommand{\xbot}{\textit{Xbot} }
\newcommand{\morphbot}{\textit{morph-bot} }
\newcommand{\frogs}{\textit{frog-bots} }
\newcommand{\frog}{\textit{frog-bot} }

\usepackage{multirow}

\definecolor{tableheader}{HTML}{CAB476}
\definecolor{tablesubheader}{rgb}{0.79,    0.698,    0.839}
\definecolor{thirdtablecolor}{rgb}{0.8706,    0.7961,    0.8941}
\usepackage{color,colortbl}

\def\scititle{
	Computational Design and Fabrication of Modular Robots with Untethered Control
}
\title{\bfseries \boldmath \scititle}
\author{
	Manas~Bhargava$^{1}$,
	Takefumi~Hiraki$^{2,3}$,
        Malina~Strugaru$^{1}$,
        Yuhan~Zhang$^{4}$,
        \and
        Michal~Piovarci$^{1,4}$,
        Chiara~Daraio$^{5}$,
        Daisuke~Iwai$^{6}$,
	Bernd~Bickel$^{1,4}$\and
        \small$^{1}$Institute of Science and Technology Austria, Klosterneuburg \& 3400, Austria. \and
	\small$^{2}$University of Tsukuba, Tsukuba\& 305-8550, Japan.\and
        \small$^{3}$Cluster Metaverse Lab, Tokyo \&141-0031, Japan.\and
        \small$^{4}$ETH Zurich, Zurich \& 8093, Switzerland.\and
        \small$^{5}$California Institute of Technology, Pasadena \& 91125, USA.\and
        \small$^{6}$The University of Osaka, Osaka \& 560-8531, Japan.\and
	\small$^\ast$Corresponding author email: manas.bhargava@ist.ac.at\and
}

\begin{document}

\maketitle

\begin{abstract} \bfseries \boldmath
Natural organisms utilize distributed actuation through their musculoskeletal systems to adapt their gait for traversing diverse terrains or to morph their bodies for varied tasks.
A longstanding challenge in robotics is to emulate this capability of natural organisms, which has motivated the development of numerous soft robotic systems. However, such systems are generally optimized for a single functionality, lack the ability to change form or function on demand, or remain tethered to bulky control systems.
To address these limitations, we present a framework for designing and controlling robots that utilize distributed actuation. We propose a novel building block that integrates 3D-printed bones with liquid crystal elastomer (LCE) muscles as lightweight actuators, enabling the modular assembly of musculoskeletal robots. We developed LCE rods that contract in response to infrared radiation, thereby providing localized, untethered control over the distributed skeletal network and producing global deformations of the robot.
To fully capitalize on the extensive design space, we introduce two computational tools: one for optimizing the robot’s skeletal graph to achieve multiple target deformations, and another for co-optimizing skeletal designs and control gaits to realize desired locomotion.
We validate our framework by constructing several robots that demonstrate complex shape morphing, diverse control schemes, and environmental adaptability.
Our system integrates advances in modular material building, untethered and distributed control, and computational design to introduce a new generation of robots that brings us closer to the capabilities of living organisms.
\end{abstract}
\paragraph*{Summary}
Inspired by vertebrates, we present active and untethered control of modular musculoskeletal robots optimized with computational tools.

\noindent

\section*{Introduction} \label{introduction}

\begin{figure}[!ht]
    \centering
    \includegraphics[width=0.96\linewidth]{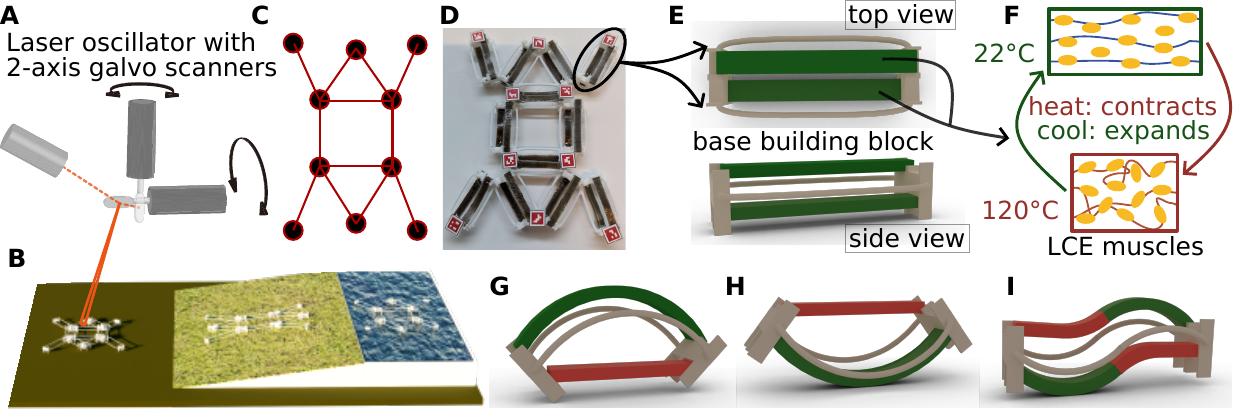}
    \caption{
    Our untethered laser heating setup (\textbf{A}) allows for remote and active actuation of robots. The robotic designs and controls are optimized for different functionalities (\textbf{B}). For every task, an optimized skeletal graph (\textbf{C}) is used for modular fabrication of the robot (\textbf{D}) with the help of base building blocks (\textbf{E}). Each building block is fabricated with a 3D-printed bone and equipped with two LCE muscles (\textbf{F}) that serve as untethered actuators. The building block can bend preferentially (\textbf{G}), (\textbf{H}) and shrink (\textbf{I}), depending on the selective activation of the muscle.
    }
    \label{fig:fig-1}
\end{figure}

Natural organisms have developed remarkable mobility to traverse challenging terrains and perform complex tasks. Their ability to readily adapt to their environment lies in their structural design; instead of specialized appendages, they leverage general building blocks shaped by evolution.
Vertebrates demonstrate this remarkable diversity, with different species having evolutionarily optimized skeletons tailored to their specific habitats.
Further enhancing this design, a distributed network of muscles empowers these skeletons for intricate motions. This synergy creates a powerful and versatile blueprint that enables vertebrates to thrive in a wide range of environments and tasks.

\begin{figure}[!ht]
    \centering
    \includegraphics[width=1\linewidth]{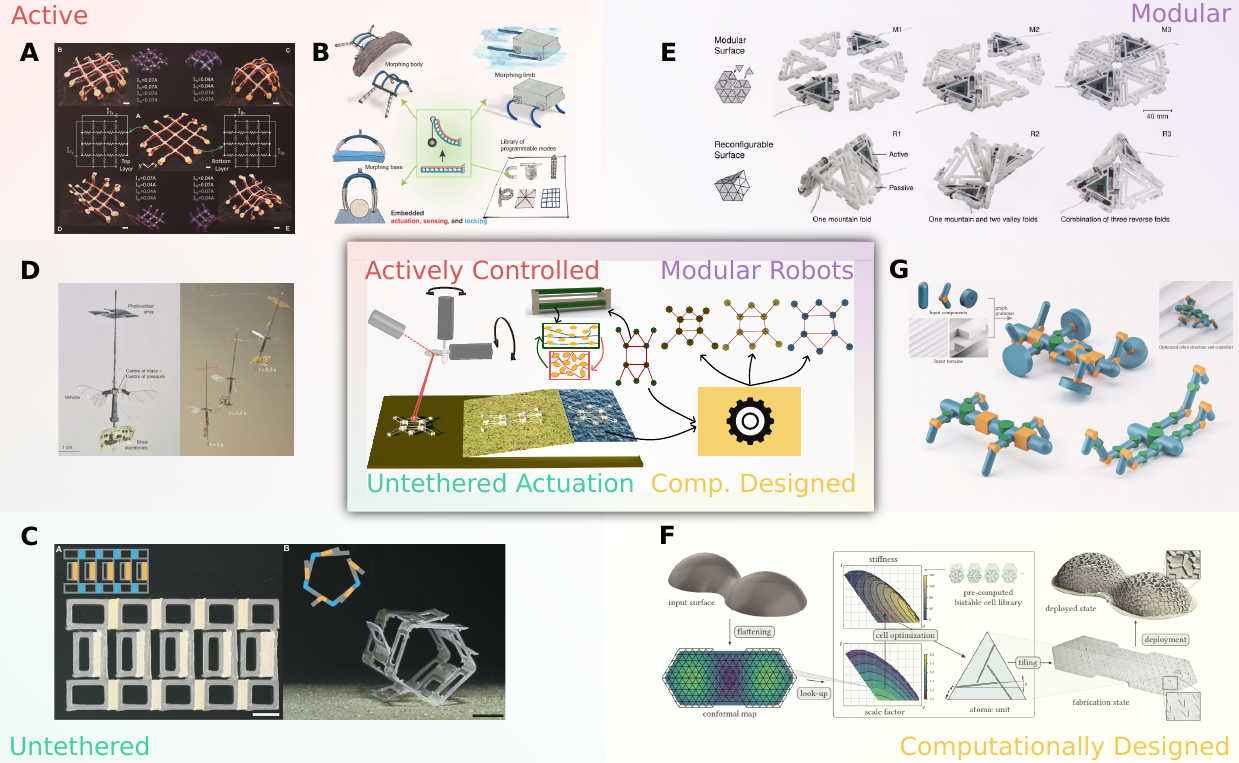}
    \caption{Active control of tethered soft robots (\textbf{A},\textbf{B}) \cite{liu2021robotic, sun2023embedded} has showcased rich performance for shape morphing and locomotion tasks. In contrast,  untethered control achieved through different passive actuation (\textbf{C}) \cite{Kotikian2019} remains limited in performance.
    Untethered robots with active control combine the best of both approaches and allow for wireless locomotion (\textbf{D}) \cite{jafferis2019untethered}. Unfortunately, such robotic systems are designed by hand and fail to scale up.
    From a design perspective, modular robots have shown promise by combining basic building blocks to perform shape-morphing or locomotion tasks (\textbf{E}) \cite{belke2017mori}. Furthermore, computational design tools allow for structural optimization of different shape morphing structures that are passively controlled (\textbf{F}) \cite{Panetta2021}. Modular robots created using computational design tools promise significant potential as demonstrated by the conceptual robots in (\textbf{G}) \cite{zhao2020robogrammar}.
    Our system uniquely integrates these different domains, enabling computational design of modular robots with active and untethered control to perform both shape morphing and locomotion tasks. }
    \label{fig:fig-2}
\end{figure}

This contrasts with the capabilities of human-designed robots that mimic natural organisms --- soft robots. Soft robots have gained traction in the robotics community due to their ability to interact safely with humans or delicate environments. They are typically made of soft elastic material, where a single spatial location can generally undergo two modes of deformation: it can either bend or shrink.
These modes of deformation have inspired many researchers to assemble individual blocks into structures that can be passively actuated into target shapes.
\cite{Konakovic2018, Chen2021} introduced an auxetic building block with a negative Poisson’s ratio that passively transforms a flat-fabricated configuration into target 3D shapes.
\cite{Panetta2019, Ren2022} utilized building blocks composed of elastic beams, rigid plates, and hinge joints to design mechanically deployable structures.
\cite{Panetta2021} introduced inflatable fusing curves as building blocks in which the flat design is encoded to passively inflate to a target 3D shape.
\cite{4DMesh2018, guseinov2020programming} used shrinkable materials as building blocks that are passively actuated with hot water to achieve the desired 3D shapes. Similar building blocks were used by \cite{Inkjet4DPrint2023} to print a heat-shrinkable base sheet that self-folds into target origami objects.
However, without active control of each individual unit, these systems remain limited in their ability to support general-purpose soft robotic applications.

On the flip side, researchers proposed robotic materials embedded with actively controlled actuators.
\cite{liu2021robotic, bai2022dynamically} designed robotic surfaces with high spatio-temporal control to perform shape morphing and object manipulation.
\cite{sun2023embedded, patel2023highly, sihite2023multi} designed shape morphing robots capable of performing locomotion tasks across a variety of terrains.
Active control of piezoelectric actuators was demonstrated by \cite{chen2023bio} to develop small soft robots capable of navigating complex environments.
Unfortunately, these robotic systems are manually designed, making it difficult to generalize their framework to broader applications.

Once the base design of the robot is fixed, a set of actuators is required to drive its motion. These actuators can be activated by external force \cite{Konakovic2018, Panetta2019, Ren2022}, pressure differentials \cite{Panetta2021, wang2023pneufab}, or by deforming the material itself \cite{guseinov2017curveups, 4DMesh2018, aharoni2018universal, guseinov2020programming, liu2021robotic,  jourdan2023shrink, patel2023highly}. Unfortunately, unlike natural organisms, machines cannot rely on distributed networks of active elements. Instead, their actuators must be individually connected to a control unit and a power source. Such a physical tethering limits the form of the robot and often restricts the machine to a single location in space, a glaring concern in the soft robotics community \cite{patel2023highly}.
To address this issue, researchers have investigated untethered control of robotic materials \cite{sabelhaus2015system, jafferis2019untethered, kim2018printing, rich2018untethered, Kotikian2019, zhai20214d, wu2022locally, lee2023magnetically, chen2023light, jung2024untethered}. However, such methods have lacked both the structural build-up of robotic surfaces using building blocks and their active control. It therefore remains an open challenge to develop an untethered actuation method that is capable of activating a large number of actuators simultaneously, required for creating modularly designed and actively controlled soft robots.

Despite existing limitations, recent advances in material science and control systems have introduced new challenges for robotic design. As the number of design options increases, identifying the most optimal designs becomes increasingly complex. The computer graphics community has developed various tools to design passively controlled deformable structures, including numerical optimization models \cite{Skouras2013, Chen2021, guseinov2017curveups, Konakovic2018, Panetta2021, Ren2022, jourdan2023shrink}, reinforcement learning methods \cite{min2019softcon, zhao2020robogrammar, park2021computational}, and more recently, differentiable predictive models \cite{ma2021diffaqua, xu2021end, gjoka2024soft}. However, the application of these tools in the design of actively controlled robots remains limited \cite{schulz2017interactive}.
We summarize these key aspects of robotic designs in \autoref{fig:fig-2}, highlighting the distinctions between modular and non-modular designs, active and passive actuation in soft robots, and tethered versus untethered control systems.

\begin{figure}[h]
    \centering
    \includegraphics[width=\linewidth]{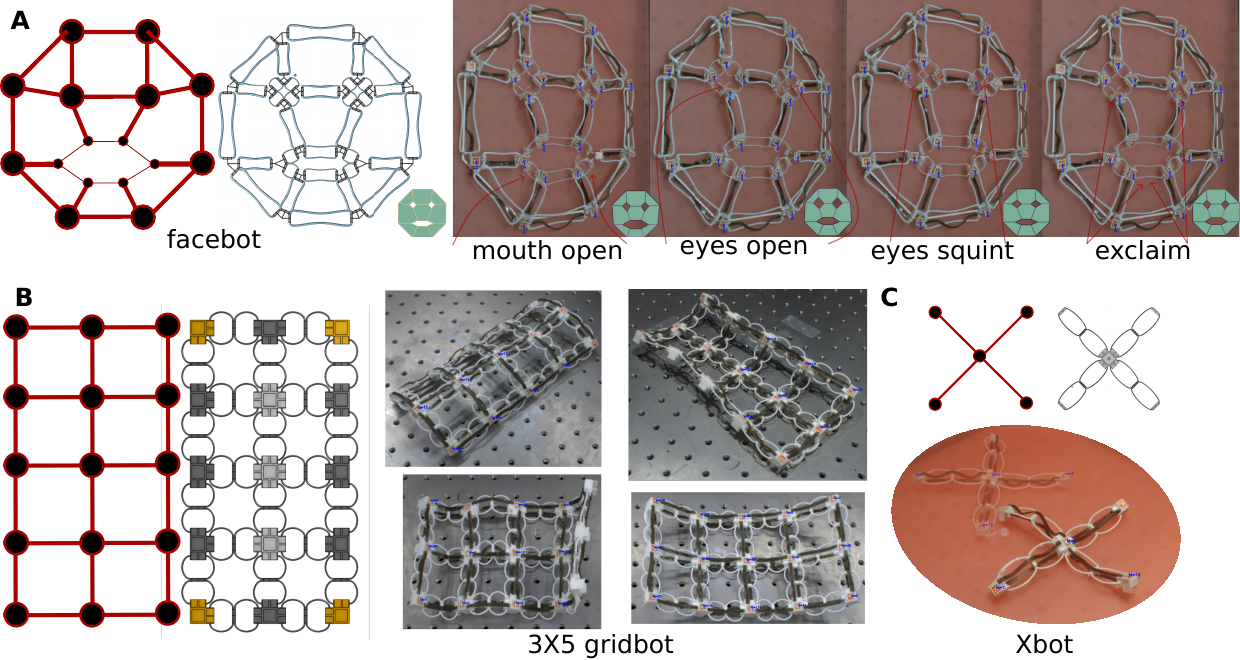}
    \caption{We illustrate our design system using \facebot (\textbf{A}), \gridbot (\textbf{B}) and \xbot (\textbf{C}). For each robotic design, we start with a graph representation of the robot, from which the CAD file is automatically generated. The CAD file is then used to 3D print the robot, after which muscles are added making the robot ready for actuation.
    \facebot is equipped with one muscle per bone that can be shrink to create different facial expressions.
    \gridbot is a $3\times5$ grid that is actuated to morph into multiple different shapes.
    \xbot is a simple X shaped robot that showcase simple locomotion mimicking a 4 legged caterpillar.
    }
    \label{fig:fig-4}
\end{figure}

In this work, we seek to address these long-standing challenges in soft robotics by taking inspiration from natural organisms. Instead of using specialized appendages or passively actuatable materials, we propose using an active actuator as the base building block (see \autoref{fig:fig-1} (E)). The robotic design is then a distributed collection of these actuators represented as edges, together forming a skeletal graph. To physically realize the robot, the building block is made from a rigid 3D-printed structure that acts as the bone. It is further equipped with two soft liquid crystal elastomer (LCE) actuators that act as muscles (see \autoref{fig:fig-1} (F)) to shrink the bone or perform preferential bending (see \autoref{fig:fig-1} (G,H,I)). To avoid tethering the individual muscles, we propose a novel activation system. We achieve this by modifying the chemical composition of the LCE muscles so they can be activated by an infrared laser, enabling active and untethered control of individual muscles and their parent bones. Finally, the collective behavior of the individual bones generates the final form and functionality of the robots.
Our proposed material and actuation system significantly broaden the space of admissible robotic designs.
To showcase our method, we first introduce three simple robots (\autoref{fig:fig-4}): a \facebot that exhibits the basic functionalities of LCEs as muscles to control the skeleton by performing different facial animations, a $3\times5$ \gridbot to show basic shape morphing capabilities from flat sheet to multiple target shapes in 3D and an \xbot that demonstrates untethered locomotion on flat terrain.

Furthermore, to fully leverage the design possibilities that our system unlocks, we equip it with two computational tools.
The first tool optimizes the skeletal design of the building blocks to create robots that can achieve a set of user-defined  target configurations. To this end, we build a \morphbot that is first verified for its functionality through simulation, and then its performance is demonstrated in the real world (\autoref{fig:fig-5}).
The second tool co-optimizes the skeletal design and its control to enable the robot to perform target functions in simulation. We use this tool to design and control different \frogs that exhibit locomotion across diverse terrains (\autoref{fig:fig-6}).
The details of the fabricated robots are summarized in \autoref{table:table-robots}.
These robots collectively illustrate the potential of our universal building block of distributed actuators for creating versatile robots.
We believe that our computational design system for the fabrication of untethered modular robots will pave the way for future soft robotic systems.

\section*{Results} \label{sec:results}

We discuss the key concepts and the design choices we made to build the robots.
\paragraph{Universal Building Block}
The building block is a single line element, as shown in \autoref{fig:fig-1}. It consists of two joints and two elastic rods that form the flexible \textit{bone} and two shrink actuators acting as \textit{muscles} that are attached to it. We 3D-print the bones (see \autoref{sec:methods} \textit{Materials and Methods: Fabrication and Characterization of 3D-printed bones}) and chemically synthesize the muscle material in the laboratory (see \autoref{sec:methods} \textit{Materials and Methods: Fabrication and Characterization of LCE-based muscles}).
The contraction of a single actuator bends the bone element with a designed directional bias, while the simultaneous contraction of both actuators results in the shrinkage of the bone element (see \autoref{fig:fig-1} (E)). We validate this experimentally in supplementary video: S1.
Our building block demonstrates an excellent motion range, achieving up to 33\% compression from its original length when shrinking, and a preferential curvature of $\pm\ 90$ degrees when bending.

\begin{figure}[!h]
    \centering
    \includegraphics[width=\linewidth]{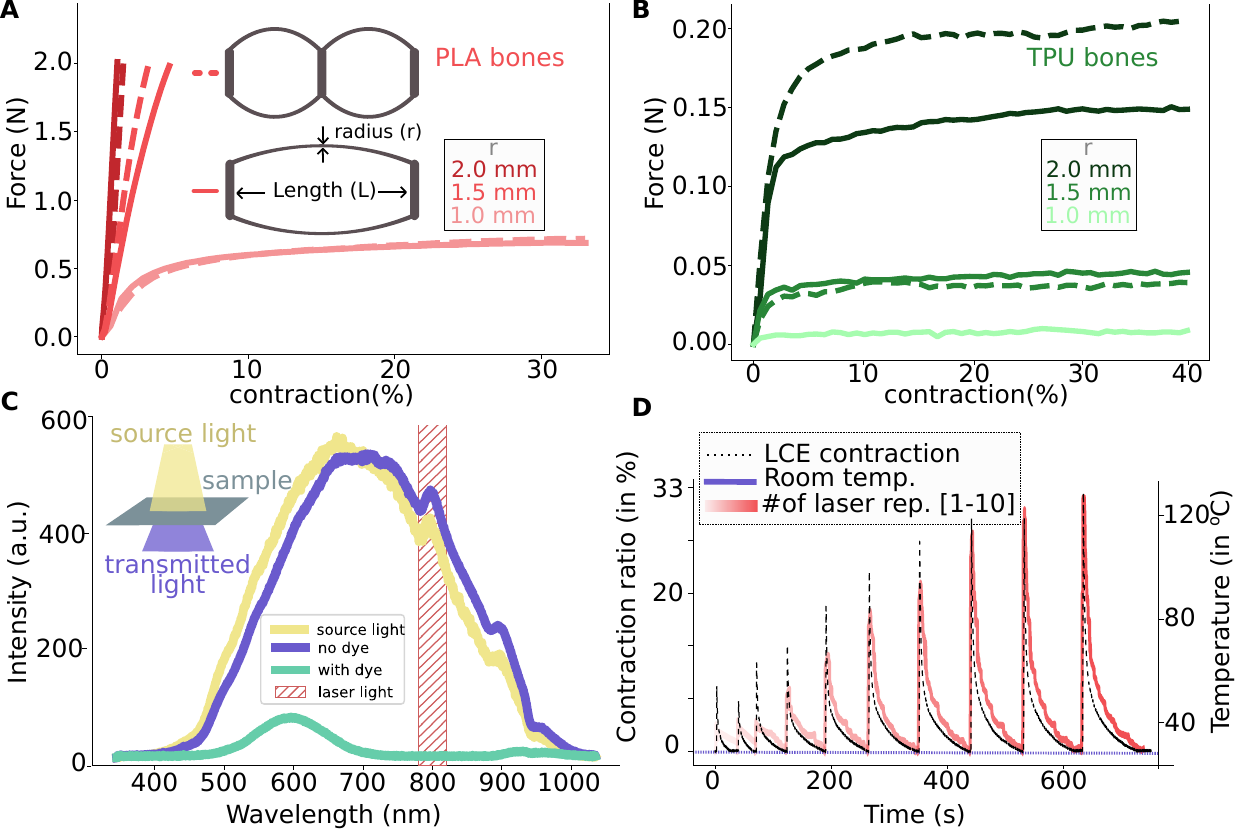}
    \caption{Material Characterization: (\textbf{A},\textbf{B}) We characterize the flexibility and bending force of our 3D-printed bones, which are influenced by the material (TPU/PLA) and the cross-section radius of the springs. (inset: 3D-printed bone with and without connectors required for shrinkage behavior of bones)
    (\textbf{C}) We demonstrate the absorption efficiency of IR820 in the infrared spectrum, showing its effectiveness in absorbing laser radiation (\textbf{D}). We characterize the response of LCE muscles by varying the number of times (1–10) the laser is swept at the highest intensity and measuring the resulting contraction, temperature changes, heating time, and cooling time. Notice that as we increase the number of laser repetitions, we are able to achieve the maximum 33\% contraction of our LCE muscles using the laser setup.}
    \label{fig:fig-3}
\end{figure}

\paragraph{Laser Heating System}
To actuate our proposed building block, we developed a custom infrared laser heating system based on previous research~\cite{hiraki2020} (see \autoref{fig:supfig-1}).
We use a high-intensity near-infrared laser diode together with a Galvoscanner.
The Galvoscanner is controlled with the help of a microcontroller.
To ensure precise laser actuation, we equip our system with six RGB cameras and perform volumetric calibration of the $35\times35\ cm^2$ experimental laser area.
Our system transmits laser instructions in packets that encode the start and end points of the trajectory, intensity, scanning speed, and total actuation time by specifying the repetition count.
We characterize the laser heating system (see \autoref{sec:characterize-laser-heating-system} \textit{Materials and Methods: Designing and characterizing the control system}) to ensure that it works in harmony with the untethered actuators introduced next.

\paragraph{Untethered Actuatable Muscles}
The actuators in our building block are rod-shaped Liquid Crystal Elastomers (LCEs). LCE muscles are heat-responsive structures that shrink in length when heated and return to their original shape when cooled to room temperature (see \autoref{fig:fig-1}).
Since the inception of LCEs in the late 1980s, the materials have attracted a significant amount of attention.
Researchers have explored various ways of synthesizing and controlling LCEs \cite{terentjev2025liquid}, ranging from tethered control \cite{yuan20173d, liu2021robotic} to untethered control \cite{Kotikian2019, zhu2024light, aharoni2018universal}. However, previous methods have had limited success in incorporating untethered and active control of LCEs, resulting in restricted design space for fabricated robots.
In our approach, we enable the use of LCEs as untethered muscles by incorporating an infrared-absorbing dye (IR820) into the material, allowing it to be heated via infrared radiation. This enables localized, untethered control of the LCE muscles using our proposed laser system.
We characterize the LCE muscles with the laser setup (see \autoref{sec:characterize-muscles-with-laser} \textit{Materials and Methods: Fabrication and Characterization of LCE based muscles},  and \autoref{fig:fig-3} (C) and (D)).

\paragraph{Design Space of Robots}
The first step in creating a robot is to construct its skeletal graph ($\mathcal{N}$, $\mathcal{E}$).
The skeletal graph consists of a collection of edges $\mathcal{E}$ that are physically realized with the building blocks, and the nodes $\mathcal{N}$ which represents the joints where the edges meet.
From this graph structure, we automatically generate the CAD model.
We fabricate the model with a Fused Deposition Modeling 3D printer.
The flexibility of the built robot depends on the flexibility of the individual bones, which in turn depends on the material used and the cross-section radius as illustrated in \autoref{fig:fig-3} (A,B) and \autoref{sec:bones}: \textit{Materials and Methods: Fabrication and Characterization of 3D-printed Bones}.
We chose the material and thickness for the robots based on the task at hand.
We report the skeletal design details for our robots in \autoref{table:table-robots}.
The maximum number of edge elements that can be actuated by our setup to achieve the target configuration is constrained by the physical limitations of the system. We explain these constraints, along with other design and fabrication aspects in the supplementary material (see \autoref{sec:appendix} \textit{Supplementary Information: Number of actuators in a robot, Generating the CAD file, 3D printing the skeleton}).

\paragraph{Control Algorithms} \label{sec:control-algorithms}
Our unique actuation system utilizes infrared radiation heating for untethered control of the LCE muscles which presents a particular set of challenges. Specifically, for any LCE muscle undergoing actuation, the instantaneous length varies non-linearly with temperature. This is explained by the material properties of the LCEs \cite{warner2007liquid} and is experimentally reported by \cite{saed2016lcefab, liu2021robotic}.
Thus, to ensure precise actuation of the LCE muscles, we place them in precisely fabricated slots on 3D-printed bones and equip them with a \textit{tracking system}. Every joint where two bones meet is equipped with an ArUco Marker that provides real-time information on the position of the bones, allowing close-loop control of the robots (details in
\autoref{sec:appendix}
\textit{Supplementary Information: Tracking ArUco Markers}).
Now, given a task at hand, we use customized algorithms to generate laser actuation profiles that provide the laser with an instruction set to scan the individual muscles of the robot.

For \textbf{shape morphing} tasks, we generate a laser actuation profile that gradually transforms the flat-fabricated robot into its target shape. Since both the flat-fabricated shape and the target shape share the same connectivity, their edges form a bijection. This allows morphing between the two, given the shrinkage and curvature information for both shapes \cite{chern2018shape}.
The shrinkage ratio is a continuous parameter that we navigate by heating the muscles gradually until they reach their target contracted length. Curvature dictates the discrete choice of which muscle to heat in a bone. A positive curvature is achieved by heating the bottom muscle, while a negative curvature is achieved by heating the top muscle. The edges are then placed in a priority queue, with priority set based on how far an edge is from the target contraction ratio. To avoid overheating any given muscle, we penalize a muscle's priority by checking the number of visits to that muscle. This heating process results in the gradual actuation of the robot until each edge reaches its target length and curvature, thus yielding the final shape.

For \textbf{locomotion} tasks, we generate laser actuation profiles that activate the muscles, resulting in their periodic motion. The amplitude, phase, and frequency of the periodic motion are tuned for each muscle by controlling the laser intensity, scan speed, and actuation time using real-time feedback on their position.

\section*{Physical Robots}
We demonstrate our proposed system for building robots using universal building blocks, coupled with untethered control by fabricating the physical prototypes listed in \autoref{table:table-robots}.
In total, we present five demos: the shape animation of \facebot\hspace{-4pt}, the shape morphing of a $3\times5$ \gridbot from flat to multiple 3D target structures, the computational design of shape-morphing \morphbot from flat fabricated into user-defined geometric shapes, simple locomotion with the \xbot and multi-terrain navigation by the \frogs\hspace{-4pt}, which are optimized for their design and control.
In the following sections, we discuss the design and control of the fabricated robots in detail.

\begin{table*}
    \begin{center}{
    \caption{Table showcases the skeletal graphs, their physical parameters and the intended tasks of the fabricated robots}.
    \label{table:table-robots}
    \resizebox{1.0\linewidth}{!}
    {
    \renewcommand{\arraystretch}{1.2} %
        \begin{tabular}{l||c|c|c|c|c|c}
        \toprule
        \rowcolor{tableheader}
        Name & Nodes ($\mathcal{N}$)  & Edges ($\mathcal{E}$) & Task & Material & Cross Sec. radius & Suppl. Video \\
        \rowcolor[HTML]{EFEFEF}\facebot & 22 & 19 & 2D Shape animation & TPU & $2\ mm$ & S2\\
        \midrule
        \rowcolor[HTML]{EFEFEF}\gridbot & 15 & 22 & $2D \rightarrow 3D$ morphing & TPU & $2\ mm$ & S3 \\
        \midrule
        \rowcolor[HTML]{EFEFEF}\xbot & 5 & 4 & 2D Locomotion & TPU & $2\ mm$ & S4\\
        \midrule
        \rowcolor[HTML]{EFEFEF}\morphbot & 15 & 33 & $2D \rightarrow 3D$ morphing & TPU & $2\ mm$ & S5\\
        \midrule
        \rowcolor[HTML]{EFEFEF}\frogs & 10 & 12 & multi-terrain navigation  & PLA & $1\ mm$ & S6\\
        \midrule
        \end{tabular}
        }
    }
    \end{center}
\end{table*}

\paragraph{Face-bot}
Activation of individual muscles is a fundamental capability exhibited by natural organisms.
Each muscle is part of the distributed actuation system and can be controlled independently, with their combined effect results in the final shape that the organisms acquires. We demonstrate this basic functionality using the \facebot (see \autoref{fig:fig-4} and supplementary video S2).
\facebot\hspace{-4pt}'s skeletal graph consists of 22 nodes and 19 bones. Each bone is equipped with one muscle placed on the bottom side. The skeleton is 3D-printed with flexible TPU material with a $2\ mm$ cross-sectional radius.
We design different face animation patterns by contracting different sets of muscles. The target lengths of the animated shapes are then fed as input to the shape-morphing control algorithm
to generate the required laser-heating profiles.
Finally, we actuate the \facebot with the laser heating profiles to obtain the desired facial expressions.

\paragraph{Grid-bot}
The \gridbot (see \autoref{fig:fig-4} and supplementary video: S3) is a $3\times5$ grid-shaped robot with its skeletal graph consisting of 15 nodes and 22 edges. The skeleton is made of flexible TPU material with a cross-sectional radius of $2\ mm$, and each bone is equipped with two muscles to allow preferential bending. We provide different target configurations as input and compute the desired length and curvature per edge. We use the shape-morphing control strategy to obtain the laser-actuation profiles and apply them to morph the \gridbot into corresponding three-dimensional shapes.

\begin{figure}[H]
    \centering
    \includegraphics[width=0.95\linewidth]{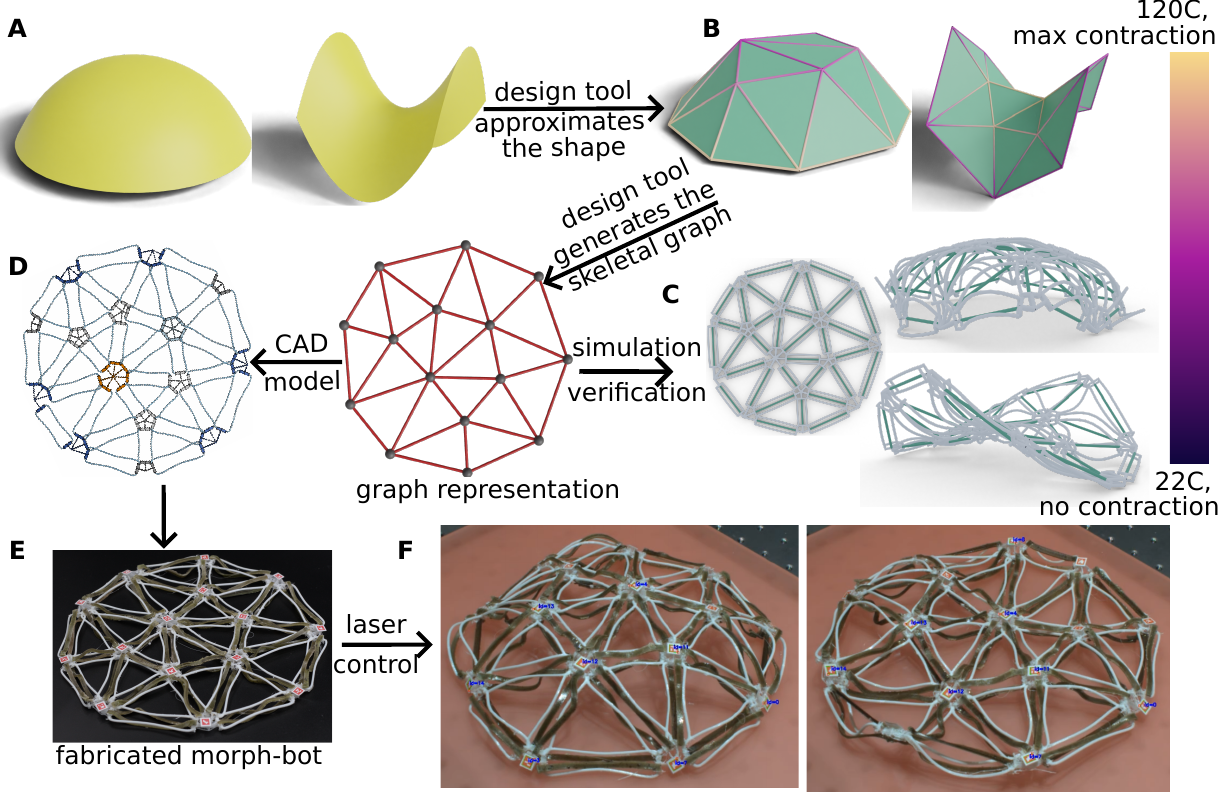}
    \caption{We demonstrate how our computational design tool is used to create shape morphing robots. Starting with user-provided input shapes (\textbf{A}) \textit{dome} and \textit{saddle}, we use a computational design tool to generate the skeletal graph. This graph tessellates the input shapes into a coarse representation (\textbf{B}) and uses the bijection between the edges to compute the shrinkage ratio and bending direction. We use the per-edge information to generate the laser actuation profile with the shape-morphing control algorithm. Before fabrication, we use a numerical simulator (\textbf{C}) to verify the design (\textbf{D}). After verification, we 3D print the robot and attach the LCE muscles (\textbf{E}). Finally, we use the laser actuation profile to morph the fabricated \morphbot towards the target 3D shapes (\textbf{F}). The system achieves the dome shape with good visual accuracy but only actuates the saddle to a partial configuration where the sharp rising edges of the desired saddle shape are countered by gravitation force.}
    \label{fig:fig-5}
\end{figure}

\paragraph{Morph-bot}

The range of shapes that we can achieve by morphing grid-like robots is inherently limited. The actuation profiles can fail to actuate the robot to its target configuration when gravitational forces are comparable to the robot’s bending energy. Moreover, it remains unclear how to design a grid-like skeletal graph that can closely approximate complex user-defined target configurations.
To this end, we develop a tool (see
\textit{Material and Methods: Computational design tool for shape morphing}) that considers the fabrication constraints of our framework and generates a skeletal graph that is optimized to approximate the user-defined target configuration when actuated using untethered laser control. Furthermore, we numerically verify the functionality of both the skeletal graph and the laser actuation profile through physics-based simulation to ensure the correct behavior of the robot after fabrication.
We use this tool to design the \morphbot\hspace{-4pt}.

\textit{Morph-bot} is provided with a \textit{Dome} and a \textit{Saddle} as target shapes, and the design tool generates its skeletal graph that contains 15 nodes and 33 edges. We use this skeletal graph to first generate the CAD model of \morphbot\hspace{-4pt}. This CAD model is used to numerically verify \morphbot with the simulation tool. Finally, we fabricate the \morphbot using flexible TPU material with a $2\ mm$ cross-section radius and equip each bone with two muscles to allow for preferential bending. We use the shape-morphing control algorithm, to morph the flat-fabricated structure to its target configurations. \autoref{fig:fig-5} and supplementary video: S5 show the \morphbot in action and provide physical validation of the computational design and simulation pipeline.
The \morphbot successfully transforms into a dome shape with good visual fidelity but achieves only a partial transformation into the saddle shape. This limitation underscores the simulation-to-reality gap as the gravitational forces dominate the TPU made \morphbot\hspace{-4pt}, preventing the structure to reach complete transformation and presents venue for future research in customized physics based simulation.

\paragraph{Xbot}
Next, we leverage the universal building block and the untethered control system to build locomotive robots. To showcase this, we built \xbot (see \autoref{fig:fig-4} and supplementary video: S4) with a skeletal graph consisting of 5 nodes and 4 bones, and equipped with two muscles per bone. To walk in a given direction, we actuate its hind legs (with respect to the direction) with a periodic wave similar to the motion of a caterpillar \cite{xu2022locomotion} using the locomotion control algorithm.

\begin{figure}[H]
    \centering
    \includegraphics[width=\linewidth]{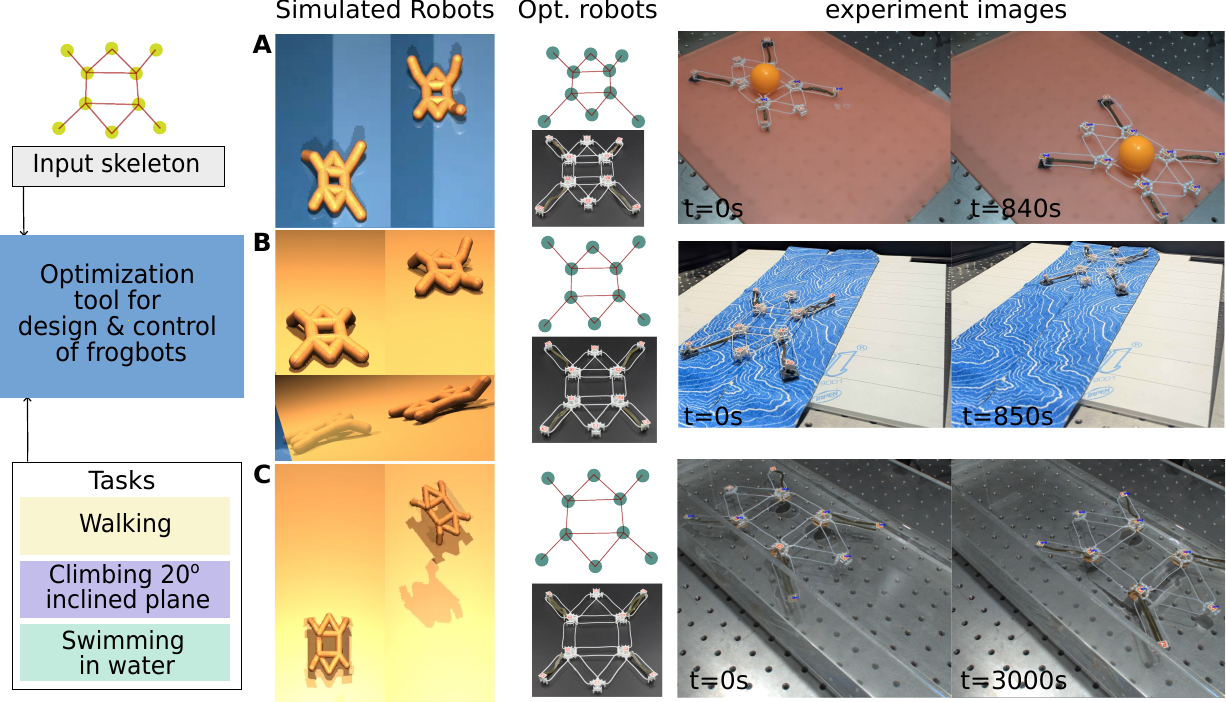}
    \caption{
        Starting with an input skeletal designs,
        we demonstrate the performance of our method by optimizing a \frog design for three different tasks: walking (\textbf{A}), climbing a 20-degree inclined plane (\textbf{B}) and swimming (\textbf{C}). \frogs were optimized for their respective terrains by learning both their skeleton and gait patterns.
        The optimized designs were then used to fabricate the \frogs and their gaits were hand-tuned for the real-world performance.
        For each terrain, we first showcase the optimized robots in the simulation. We then showcase their optimized skeleton from which the final fabricated model is obtained.  Finally, we showcase the real world performance by taking their start and end snapshots.
    }
    \label{fig:fig-6}
\end{figure}

\paragraph{Frog-bots}
Finally, to showcase the vast design and control space offered by our universal building block and untethered control system, we introduce multi-terrain navigating \frogs that can walk on a flat surface, climb an inclined plane, and swim in water. We start with a skeletal graph of 10 nodes and 12 bones and fabricate a standard \frog that can navigate the three terrains. We then use our design tool (see \autoref{sec:dct}:
\textit{Materials and Methods: Exploring the design and control of terrain navigating robots}) to explore the skeletal design and control gaits and create three optimized \frogs\hspace{-4pt}, one for each terrain.
The fabricated \frogs use rigid PLA material of $1\ mm$ cross-sectional radius and the locomotion control algorithm is fine-tuned for their specific terrains (see \autoref{fig:fig-6} and supplementary video: S6).

\section*{Discussion}
To conclude, we present a novel system that enables the structural fabrication of soft robots by combining our universal building blocks with a distributed, untethered actuation system. The fabricated robots showcase complex motions ranging from shape morphing to locomotion, mimicking the natural organisms.
Our proposed system opens up a new optimization paradigm for robotic designs that enables us to explore both the degrees of freedom in design and control jointly. We present our computational design tools that utilize these degrees of freedom to build physical robots with applications in multi-shape morphing and diverse terrain navigation.
We envision a future where our system can be used by both researchers and engineers to deploy robots in the real world to solve practical tasks.

Our current framework lies in an experimental laboratory setting, which constraints the structural tessellation of the fabricated robots and restricts the range of their movement.
We believe that advances in material science for 3D printing bones \cite{buchner2023vision} and the synthesis of LCE-based muscles \cite{zang20244d} along with developments in high-intensity laser actuation would help streamline the fabrication of a robot with hundreds of actuators.
Engineering support that harnesses the sun’s infrared radiation could enable control of robots beyond the experimental setting \cite{clery2022has}.
Our robots provide a platform for the machine learning and computational design community to expand upon by providing enhanced simulation tools that reduce the sim-to-real gap and equip them with learning-based methods \cite{sun2024machine, Genesis} to design the next generation of robots.
We envision that our system will help the community make a fresh leap into a new paradigm of computationally designed modular soft robots, optimized for task-specific specifications by drawing inspiration from our proposed building block with active and untethered control.

\section*{Materials and Methods} \label{sec:methods}
\paragraph{Fabrication and Characterization of 3D-printed Bones} \label{sec:bones}
We 3D print the skeleton of our robots by assembling individual bones. Each bone is parameterized with length as a degree of freedom. The material used for printing, as well as the cross-sectional thickness, determine the flexibility of the bone (see  \textit{Supplementary Information: 3D printing the skeleton}). We characterize these bones using the ZwickRoell tensile test machine. We perform a compression test to measure the force needed to bend each bone. The loading threshold is set either to the maximum contraction length achievable by LCEs (33\%) or to a force limit of $2\ N$ (twice the force LCE muscles can apply).
\autoref{fig:fig-3} (A,B) shows the force-contraction curves for bones made of two 3D-printable materials, PLA and TPU. For each material, we analyze structures with a middle connector and without it, and three values of rod thickness.  We use $1.0\ mm$ cross section for the PLA skeleton and $2.0\ mm$ cross section thickness for the TPU skeleton for the printed robots, as they allow preferable stiffness-to-flexibility behavior.
Bones made with PLA exhibit only preferential bending due to its stiff material response, while bones fabricated with TPU showcase both preferential bending and shrinkage (see supplementary video S1)

\paragraph{Fabrication and Characterization of LCE based Muscles} \label{sec:characterize-muscles-with-laser}
LCEs exhibit reversible deformation when heated, as a result of the rearrangement of liquid crystal molecules \cite{warner2007liquid}. We fabricate rod-shaped LCEs based on the synthesis reaction reported in \cite{saed2016lcefab}. We enable the actuation of the muscles with electromagnetic radiations by incorporating an infrared-absorbing dye (IR820) within the polymer mixture.
We conducted an infrared spectrometry test to measure the IR radiation absorption of the LCE muscles. \autoref{fig:fig-3} (C) shows an absorption peak around $\lambda$ = $800\ nm$, matching the operational wavelength of the laser and enabling the efficient transfer of heat to the LCEs. The material rods achieve a maximum contraction of 33\% along the alignment direction when heated from room temperature to 120\degree C.
We provide the synthesis protocol for the LCE muscles, their dimensions, fabrication challenges and limitations in the supplementary material.

\paragraph{Assembling the Musculoskeletal Robot} \label{sec:mms}
Once the skeleton (3D-printed) and the muscles (synthesized LCEs) are ready, we attach the muscles to the skeleton using \textit{Kraftprotz} super glue. An  \mbox{$8\times8\ mm^2$} ArUco marker is placed at each joint where two bones meet.
This helps in uniquely identifying the joint.
The LCE muscles are cut with 5\% extra length as slack and then glued to the endpoints at their designated location. For each edge, a bottom muscle is always on the left side of the edge and the top muscle on the right side, with the direction guided from the lower to the higher vertex index.
This gives us an exact geometric location of the muscles based on the position and normal values of the markers on both sides of the edge. This process is repeated for all the edges and the final robot is then ready to be used with the control system (see \autoref{fig:fig-1} (d-e-f) : skeletal graph to CAD model to fabricated robot).

\paragraph{Designing and Characterizing the Control System}\label{sec:characterize-laser-heating-system}
To build our own laser heating setup, we used a laser diode system ($808\ nm$, $50\ W$) manufactured by Civil Laser. The laser beam is focused using a collimator and directed to a Galvoscanner (JD2808, Sino Galvo) through a microcontroller (Teensy 4), which controls the scanner at high speeds using the $XY2-100$ protocol. The microcontroller processes target position data to control both the scanner and laser power. The detailed calibration process for the heating system is explained in \autoref{sec:laser-calibration}: \textit{Supplementary Information: Calibrating the laser heating system}.
The laser control system with all its components is showcased in \autoref{fig:supfig-1}.

We implement a pulse-width-modulation algorithm on the Teensy-4 microcontroller to interpolate the intensity value in the range $[0-1]$ using the maximum operational current of $5\ A$ and the minimum packet communication latency of $50\ ms$.
Since most of the control actions rely on shrinking the muscles to a target length, we chose the configuration of laser parameters that allow us to actuate the maximum number of muscles concurrently. The experiment details are explained in \autoref{sec:heating-budget}: \textit{Supplementary Information: Constraints on the Number of Actuators in a Robot}.
With this configuration, we can achieve a total muscle contraction length of up to $300\ mm$. For a standard muscle of length $30\ mm$, this corresponds to contracting approximately 30 muscles, each by up to 33\%.
This total shrink length and contraction ratio define the fabrication limitations of the skeletal designs imposed by the control system (see \autoref{eq:fab-constraints}).
For this configuration, we further characterize muscle shrinkage by varying the number of repetitions of laser actuation on an LCE muscle of length $30\ mm$. \autoref{fig:fig-3} (D) showcases the shrinkage behavior of muscles when activated with the laser setup for maximum intensity while varying the number of repetitions.

\begin{algorithm}
    \caption{Pseudo Code for computational design for shape morphing structures  }\label{algorithm:pseudo-code-cds}
    \begin{algorithmic}[1]
    \State Input: 3D manifold meshes $\{\Omega^1_0(V^1_0,F^1_0)$, $\Omega^2_0(V^2_0,F^2_0)$... $\Omega^n_0(V^n_0,F^n_0)\}$
    \State Output: 3D approximated meshes $\{\Omega^{1\star}(V^{1\star},T^\star), \Omega^{2\star}(V^{2\star},T^\star)... \Omega^{n\star}(V^{n\star},T^\star)$\}]
    and 2D mesh $\Gamma^\star(V^{flat\star},T^\star)$, all sharing the same triangulation $T^\star$.
    \State Energy Function: for the current meshes $\{\Omega^i(V^i, T)\}$, $\Gamma(V^{flat}, T)$,
    \begin{flalign} \label{eq:energy-function}
        \mathcal{E} = \frac{1}{{N\choose 2}}\sum_{ij}E_{map}(\Omega^{i}, \Omega^{j})\ + \   \frac{1}{N} \sum_i E_{mesh}(\Omega^{i }) \ + \  \frac{1}{N}\sum_i E_{fab} (\Gamma, \Omega^{i }). \quad
    \end{flalign}
    Constraints: $\forall i \in [1,n]$ and $\forall ej \in T$, and $\mathcal{L}$ be the length of edge $e_j$,
    \begin{gather}
        \forall i
        \left(
        \sum_{e_j \in T} \left|
                \mathcal{L} \left( \Gamma(e_j)\right) -
                \mathcal{L} \left( \Omega^{i}(e_j)\right)
            \right|  <= 300
        \right), \notag \\
        \forall i
        \left(
        \forall_{e_j\in T}:  \frac{\mathcal{L} \left( \Gamma(e_j)\right)-\mathcal{L} \left( \Omega^{i}(e_j)\right)}{\mathcal{L} \left( \Gamma(e_j)\right)} < 33\%
        \right), \notag\\
        \forall_{e_j\in T}:  \mathcal{L} \left( \Gamma(e_j)\right) > 30 . \label{eq:fab-constraints}
    \end{gather}
    \State
    Initialize a coarse triangulation $T$ (two triangles) and then initialize $\Omega^i(V^i, T$) by selecting extreme boundary vertices of $\Omega^i_0$ as $V^i$. Initialize $\Gamma(V^{flat0}, T)$ with $V^{flat0}$ as vertices on the boundary of unit circle.
    \State Optimization: The common triangulation $T$ and the vertex positions $\{V^i\}$ and $V^{flat}$ are optimized by minimizing the energy function (\autoref{eq:energy-function}) while adhering to the the fabrication constraints (\autoref{eq:fab-constraints})
    \end{algorithmic}
\end{algorithm}

\paragraph{Computational Design Tool for Shape Morphing} \label{sec:cds}
The design tool aims to generate the skeletal graph for robots that can morph between different target deformations. It takes the desired target shapes as input and generates a flat skeletal graph, which after fabrication, can be actuated by the laser heating system to closely approximate the input target shapes.
To obtain the skeletal graph, we approximate the target shapes with the same connectivity and ensure their edge lengths are within the shrinkable range of the flat rest graph. This allows our edge-shrinkage-based method to morph between them.
We adapt the framework proposed by \cite{schmidt_surface_2023} aimed at computing surface homeomorphisms by optimizing compatible triangulations for various objects. This triangulation when embedded in 2D yields a planar skeletal graph structure.
We summarize our method in \autoref{algorithm:pseudo-code-cds}.
The detailed modifications made to the formulation proposed by \cite{schmidt_surface_2023} and the optimization steps are explained in
\textit{Supplementary Information:  Algorithmic Details of Computational Design for Shape Morphing Structures}.
We show the efficacy of our tool by building the \morphbot (see \autoref{fig:fig-5}), which morphs into target saddle and dome shapes with good visual accuracy.
We further showcase futuristic applications that the computational design tool unlocks by fabricating finely tessellated flat-fabricated surfaces that morph into multiple target configurations (see \autoref{fig:supfig-2}).

\paragraph{Numerical simulation to Verify Shape Morphing Robots} \label{sec:simulating-robots}
The numerical simulation uses the robot's skeletal graph as input.
We model the bones of the skeletal graph using discrete elastic rods \cite{bergou2008discrete}. The LCE muscles are modeled as spring actuators in the simulation, with their rest length being a function of temperature. We adjusted the material parameters of the simulation model to ensure that the motion of the base building block matches its behavior in the real world. This is done by tuning the simulation's material parameters as described in \cite{perez2015design}.
We mimic the muscle shrinkage as result of laser actuation, by gradually changing the rest length of the spring actuators from their initial length to the target length during the simulation.
Finally, we use a quasi-static simulation framework with contact and gravity to obtain the final shape of the robots after morphing.
This allows us to verify the efficacy of the shape-morphing robots before they are manufactured. \autoref{fig:fig-5} (C) shows the simulation results of \morphbot\hspace{-4pt}.

\paragraph{Exploring the Design and Control of Terrain Navigating Robots} \label{sec:dct}
We introduce a design tool to fine-tune a given skeletal robot to achieve a desired functional task. The design space includes the length of active actuators and their control policy. We jointly optimize these two spaces to obtain a skeleton with a control scheme that performs the target function. If multiple functions are desired, we output one unique optimized skeleton and a control scheme for each function.
We model the robots as rigid articulated robots and simulated them using ~\textsc{Mujoco} \cite{mujoco}.
We simulate three different target behaviors: walking on flat terrain, climbing on an inclined plane, and swimming in water.
Since ~\textsc{Mujoco} allows for fast simulation of rigid robots, we can optimize the nodal positions of the skeleton and control schemes for different target behavior using evolutionary search based algorithms.
The implementation details of our optimization framework are described in
\textit{Supplementary Information: Algorithmic Details for exploring the Design and Control of Terrain Navigating Robots}.
The optimized skeletons exhibit the highest fitness scores for their respective target behavior.
We use this tool for the designs of \frogs\hspace{-4pt}.
Modeling the robots with rigid linkages introduces a significant sim-to-real gap, leading to discrepancy between simulated and fabricated motions.
We use the skeletal design produced by our tool and manually tune the actuation profiles for the \frogs to maximize their performance in the real world.
Despite this, the design tool successfully optimizes the skeletal graph of \frogs for their specific tasks, outperforming the non-optimized skeleton \frog\hspace{-4pt}. \autoref{fig:fig-6} and supplementary video: S6, showcase the \frogs designed using this tool.

\bibliography{sn-bibliography}
\bibliographystyle{sciencemag}

\section*{Acknowledgments}
The authors express gratitude to Magali Lorion for assisting in the initial fabrication of LCEs, Pengbin Tang for providing the code for simulating discrete elastic rods, the Imaging and Optics Facility at ISTA for assisting with the spectrometry measurements, and the MIBA machine shop at ISTA for their support in manufacturing various devices. %
\paragraph*{Funding:}
This project was supported by the European Research Council (ERC) under the European Union’s Horizon 2020 research and innovation program (Grant Agreement No. 715767 -– MATERIALIZABLE). %
\paragraph*{Author contributions:}
M.B. conducted the primary research for this study, including the synthesis of LCE material, the characterization of the 3D-printed bone structures, and the characterization of the laser heating setup with LCE muscles. M.B. also designed, fabricated, and executed experiments for the various robotic systems presented; automated the CAD design pipeline; and developed the design and simulation tools described in the paper. M.B. was responsible for drafting the original manuscript and preparing the figures. Additionally, M.B. contributed to the building and calibration of the laser setup.
T.H. designed, built, and calibrated the laser actuation hardware and software and implemented the marker tracking system. T.H. also assisted in the design of the laser-based experiments.
M.S. advanced the research on LCE muscles by developing the IR-absorbing LCE materials used in the study, and conducted experiments for the mechanical characterization of the bones and muscles. M.B and M.S. designed the building block of the robots and the experimental environments for the locomotive robots.
Y.Z. contributed to the computational design tool for shape morphign structures.
M.P. contributed to the conceptualization of the design and simulation tools.
B.B. conceived the study. C.D. served in an advisory role and helped with conceptualization. D.I. and B.B. supervised the project.
All authors contributed to project discussions and the editing of the manuscript. %
\paragraph*{Competing interests:}
The authors declare that they have no competing interests. %
\paragraph*{Data and materials availability:}
The code for numerical simulations has been deposited in the
GitHub repository: \texttt{https://github.com/manas-avi/CDFMRUC}.

\renewcommand{\thefigure}{S\arabic{figure}}
\renewcommand{\thetable}{S\arabic{table}}
\renewcommand{\theequation}{S\arabic{equation}}
\renewcommand{\thepage}{S\arabic{page}}
\setcounter{figure}{0}
\setcounter{table}{0}
\setcounter{equation}{0}
\setcounter{page}{1}

\appendix

\section*{Supplementary Information} \label{sec:appendix}

\paragraph{LCEs Synthesis Protocol} \label{sec:lce-synth}
\sloppypar{}
RM257 (monomer, CAS:174063-87-7, 12 g, 20.39 mmol, $\>98\%$ purity) and 2-Hydroxy-4'-(2-hydroxyethoxy)-$2$-methylpropiophenone (photoinitiator,
CAS:106797-53-9, 0.077 g, 0.34 mmol, 98\% purity) were dissolved in toluene (5.55 ml) at elevated temperature (80 C) with stirring. The solution was allowed to cool at room temperature after which 2,2'-(ethylenedioxy) diethanethiol (spacer, CAS:14970-87-7 2.5 ml, 15.36 mmol, 95\% purity) was then added followed by pentaerythritol tetrakis (3-mercaptopropionate) (cross-linker, CAS: 7575-23-7, 0.57 ml, 1.49 mmol, $>95\%$ purity). IR820 (IR sensitizer, 172616-80-7, 80 mg, 94.17 mmol, 100\% purity) was then added. The solution was then sonicated to ensure thorough mixing of all components. Dipropylamine (1.97 ml, 2.57 vol.\% in toluene) was then added and the resulting solution was then mixed. The solution was degassed under vacuum (45 s, 508 mm Hg) and poured into a High Density Polyethylene mold. The samples were then left to set and solidify (8 h) under ambient conditions. The solidification was then completed under vacuum (80 C, 8 h). The samples were then mechanically stretched (150\% of the original length) and cured (365 nm, 9 W, 30 min each side) to fix the alignment of the mesogens.
\paragraph{Dimensions of Standard LCEs} \label{sec:LCE-dimensions}
The maximum contraction, the required laser actuation, and the actuation strength of the LCEs rods are functions of their cross-sectional dimensions.
LCEs were fabricated with varying widths and thicknesses, and through empirical evaluation, $5\ mm$ width and $2\ mm$ thickness were obtained as optimal dimensions. These dimensions exhibited the minimal fixity by the LCE rods, good contraction capability ($33 \%$) and provided reliable actuation behavior.
A standard LCE rod is synthesized in a mold of dimension $60\times5\times2\ mm^3$.
The synthesized LCEs were mechanically stretched to $90\ mm$ and subsequently UV-cured, retaining this extended length at room temperature. Upon heating to 120\degree C they contract back to $60\ mm$, achieving a contraction ratio of $33\%$.
Each LCE rod of length $90\ mm$  after fabrication weighs approximately $0.5\ g$ and is capable of lifting $100\ g$ of mass when heated to $120 \degree C$. This corresponds to the LCE muscle generating a force of roughly $1\ N$ while maintaining its contracted length.
Each muscle is cut to a specific length, according to the corresponding edge length in the robot's skeletal graph.
\paragraph{Limitations in using LCEs as Muscles} \label{sec:LCE-limitations}
LCE synthesis is performed under experimental conditions, which can lead to fabrication discrepancies observable in the behavior of the LCE muscles. They may not exhibit the same response when actuated with a laser. This inconsistency is primarily due to the non-uniform distribution of the IR$820$ dye, which results in clumped and uneven heating under infrared laser activation.
To address this, a closed-loop control system is implemented to monitor each bone and its corresponding muscles, adjusting the actuation profile based on the current state.
Furthermore, the laser system has a finite capacity and restricts the number of LCEs it can simultaneously heat. We explain this in the subsequent section.
Moreover, since the LCEs are glued onto the bones, a few millimeters of each LCE become nonresponsive. This imposes a minimum LCE length of $30\ mm$, which is required for reliable deformation. LCEs fabricated below this length fail to consistently achieve the desired maximum contraction of $33\%$. The minimum LCE length of $30\ mm$ is used as a fabrication constraint (\autoref{eq:fab-constraints}) in the design tool.

\paragraph{Constraints on the Number of Actuators in a Robot}\label{sec:heating-budget}
A key constraint in any robotic design is the number of actuators that can be actuated concurrently.
This limitation in our system is dictated by both the physical characteristics of the LCE muscles and the performance constraints of the untethered laser actuation system.
To estimate the number of actuators that can be admitted by our system, we conduct the following experiment.
We actuate a LCE muscle of length $30\ mm$ with the laser setup. We generate different experimental configurations ($Ei$) by varying laser speed and the number of repetitions. For each experimental configuration, we measure the operation time ($Oi_t$) of the laser to activate the LCE muscle and the cooling time ($Ci_t$) for the LCE muscle before it comes back to its pre-heated configuration. (see \autoref{fig:fig-3} (D)).
During the cooling period, the laser is available to actuate other LCE muscles. We estimate the number of LCE muscles that can be actuated in parallel under configuration $Ei$ as
\begin{equation}
N(Ei) = \frac{Ci_t}{Oi_t}. \notag
\end{equation}
We then select the configuration that maximizes $N(Ei)$ and obtain the maximum value as $30$ actuators.
Therefore, under the current setup, we can simultaneously heat 30 LCE muscles of length $30\ mm$ each,  with (maximum) contraction rate of $33\%$,
giving us a total contracted length of $30 \times 30\ mm \times 33\% = 300\ mm$.
Most robotic designs, however, do not require full contraction of each edge, but partial contraction is often sufficient to achieve the desired shape or motion.
Thus, for an edge $e_i$, with an initial length $l(e_i)$ and a desired target length $d(e_i)$, we obtain the actuation budget as the following constraint:
\begin{equation}
\sum_i^{|E|} ||(l(e_i)-d(e_i))||  <= 300\ mm .\notag
\end{equation}
This inequality serves as the feasibility constraint for any robot design and is incorporated into the design tool (see \autoref{eq:fab-constraints}).

\paragraph{Generating the CAD File} \label{sec:autocad}
We use the skeletal graph structure to generate the CAD file for the robot. The online CAD software Onshape \cite{Onshape} was used for this task.
Each node is modeled using its vertex degree and the angle between the neighboring edges that are connected to this vertex.
The position of the LCE slots is optimized using a Mathematica script \cite{Mathematica} to ensure minimum spatial usage by the nodes, since they do not contribute to actuation.
To model the edges, we use connectors with a circular cross-section to connect the two nodes.
If the edge has to be shrunk, an additional block part is generated in the middle to glue the LCEs.
Finally, all the individual nodes and bones are generated using our script and are assembled in the OnShape assembly tool following the skeletal graph's connectivity. The final CAD model is then ready to be used for 3D printing or simulation.

\paragraph{3D Printing the Skeleton} \label{sec:3dprint-process}

We used an Ultimaker S7, a Fused Deposition Modeling 3D printer to manufacture the skeletons. The 3D printer was used with Ultimaker's TPU-95A and Ultimaker's PLA (Pearl White) materials. Ultimaker's Natural PVA was used as support material.
3D prints made with TPU material are elastic, allowing for both shrinkage and preferential bending of the bones. However, skeletons made from TPU are prone to sagging under gravity. Bones printed with PLA are rigid and can only perform preferential bending, but their structural stiffness enables them to maintain geometric constraints regardless of gravitational effects. Moreover, TPU is tricky to print in combination with the support material, making the overall print process quite cumbersome, whereas printing with PLA is relatively hassle free.
Additionally, the \frog robot, was printed using the Bambu Lab X1C commercial printer with their PLA Basic material, which provides a similar mechanical response to Ultimaker PLA Pearl White, but the 3D-printed skeletons were more durable.

\paragraph{Tracking ArUco Markers}\label{sec:aruco-tracking}
We employed the AprilTag 16h5 dictionary, which can handle 30 unique IDs, enabling us to track multiple markers simultaneously.
These ArUco markers are tracked by six RGB cameras (MC031CG-SY-UB, Ximea).
To prevent imaging interference during laser operation, each RGB camera is equipped with an ND filter (NENIR260B, THORLABS) that blocks the infrared radiation emitted by the laser.
Furthermore, to avoid burning the markers with the laser, the marker is printed in red on a white background.
We used a standard corner detection and point triangulation functions from the OpenCV library to obtain the 3D location of the center and the plane normal of the markers. With our stereo-camera tracking, we can determine the 3D coordinates of our marker with up to $2\ mm$ accuracy.
The current robotic designs use fewer than 30 nodes in the skeletal graph. If more nodes are required, AprilTag 36h11 dictionary can be used.

\begin{figure}[h]
    \centering
    \includegraphics[width=\linewidth]{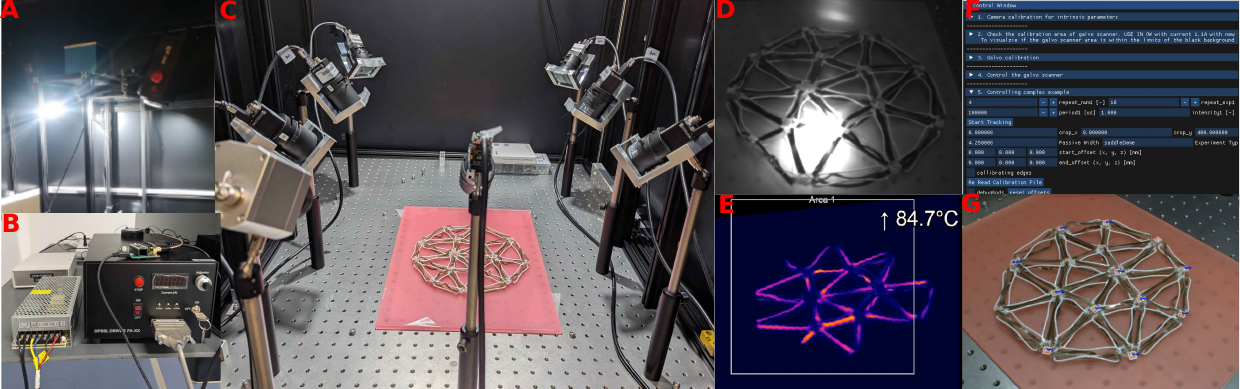}
    \caption{ (\textbf{A}) High lumen light for proper illumination, Infrared Laser and the 2-axis galvanoscanner setup. (\textbf{B}) Teensy-4 and the power supply unit for the laser. (\textbf{C}) The laser calibration and tracking setup comprising of 6 RGB, 1 monochrome and 1 thermographic camera. \morphbot visualized with monochrome camera (\textbf{D}), thermographic camera (\textbf{E}) and RGB camera (\textbf{G}). Control window in (\textbf{F}).
    }\label{fig:supfig-1}
\end{figure}

\begin{figure}[h]
    \centering
    \includegraphics[width=\linewidth]{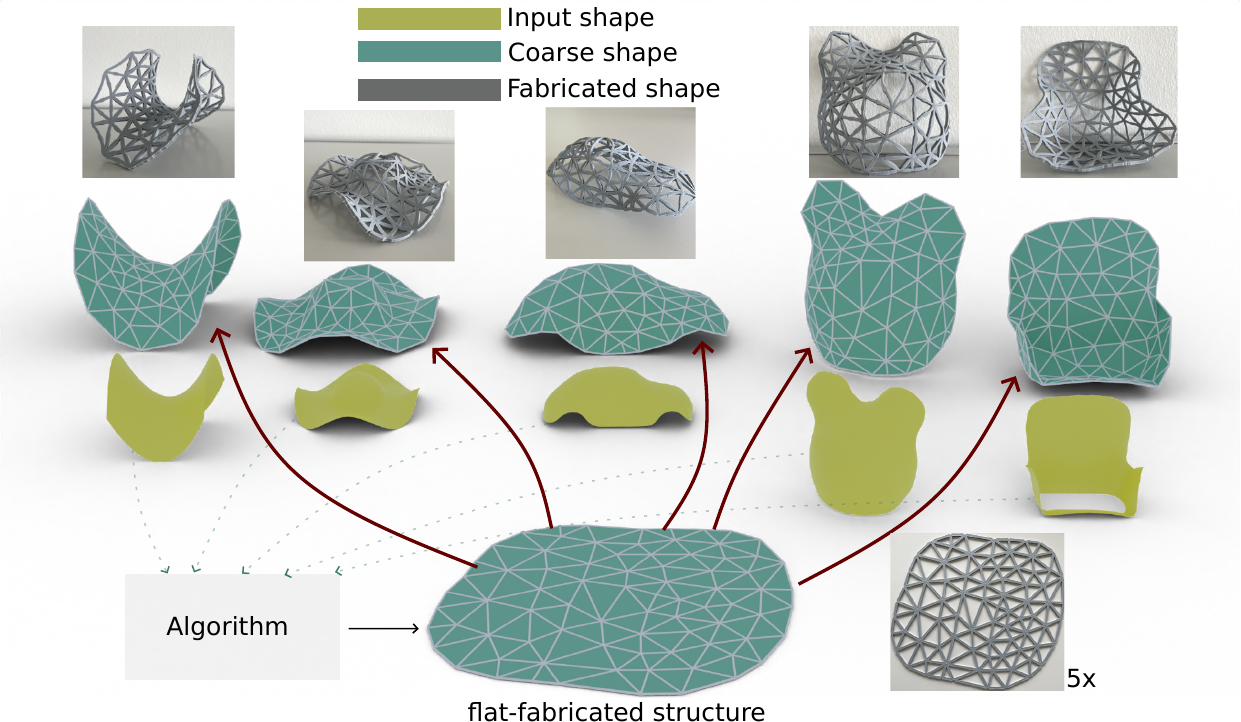}
    \caption{Application of our computational design tool (\autoref{sec:cds} \textit{Materials and Methods: Computational Design for Shape Morphing Structures}) to obtain highly-tessellated shape morphing structures. Our algorithm takes in multiple input shapes as input (in light-green) and generates a common triangulation for the skeletal graph of the flat-fabricated structure and the approximated target shapes (in dark-green). We use the \cite{4DMesh2018} framework to 3D print five  flat-fabricated structures that are visually the same but each is encoded with different shrinkage ratio per edge. These different flat-fabricated structures are passively actuated by heating in hot water and deform to their target configuration (in gray) closely approximating the input shapes.
    }
    \label{fig:supfig-2}
\end{figure}

\paragraph{Calibrating the Laser Heating System}\label{sec:laser-calibration}

First, the mirrors are moved incrementally at fixed angular intervals, and the equation of the laser line is estimated for each mirror angle. This is done by projecting the laser onto checkerboard patterns placed at known positions along the z-axis of the world coordinate system, ranging from $z = 0$ mm to $z = 40$ mm. By observing the laser’s intersections with the checkerboards, we derive the equations of the laser lines. This process constitutes the offline calibration phase. In the subsequent online phase, when a target position $(x, y, z)$ is given, the system first computes the intersection point of the laser line and the z-plane corresponding to the given z-coordinate, using the laser line equations obtained in the offline phase. Then, using Thin Plate Spline (TPS) interpolation, the mirror angles required to direct the laser to the desired position $(x, y, z)$ are estimated from the angles corresponding to these previously calculated intersection points.

\paragraph{Calibrating the Robot}\label{sec:calibrating-the-robot}
Once the robot is assembled, the next step is to calibrate the robot to remove any systematic error introduced during the manual assembly process. We calibrate both muscles of each bone by manually adjusting the marker positions to ensure the laser targets the center line of each muscle. These manual offsets are stored on the disk and later applied during the robot's control to the target function.

\paragraph{Algorithm Details of Computational Design for Shape Morphing Structures} \label{sec:details-cds}
The design tool takes as input $N$ different three-dimensional manifold meshes $\{\Omega^1_0(V^1_0,F^1_0)$, $\Omega^2_0(V^2_0,F^2_0)$... $\Omega^n_0(V^n_0,F^n_0)\}$
and output $N$ corresponding approximated meshes in 3D $\{\Omega^{1\star}(V^{1\star},T^\star)$, $\Omega^{2\star}(V^{2\star},T^\star)$... $\Omega^{n\star}(V^{n\star},T^\star)\}$ and a two-dimensional flat mesh $\Gamma^\star (V^{flat\star}, T^\star)$, all having the same connectivity $T^\star$.
Mesh $\Omega^i$ represents the current approximated mesh of the input mesh $\Omega^i_0$, and $\Gamma$ represents the current flat mesh during the optimization process. Below we describe the details and rationale behind different energy functions, initialization and optimization process used by our design tool.
\par
\textbf{Initialization}: We normalize the input meshes $\Omega_0^i$ to have unit surface area, then obtain embeddings of the normalized input meshes as the conformal parametrization map ($f$) and constrain them to lie on a unit disk using the method proposed by \cite{sawhney2017boundary}.
The  inverse of this map $f^{-1}$ is used to lift vertices in the embedding space to the input mesh using the barycentric coordinates.
We initialize the triangulation $T$ with two triangles and use this triangulation to remesh the original embeddings and initialize the 2D mesh $\Gamma$. The vertices of the remeshed embeddings and $\Gamma$ are chosen as points on the unit circle that maximize the area of the triangles.
Although we define the energies below with respect to the approximated meshes ${\Omega^i}$, the actual optimization variables are the remeshed embeddings. The approximated surfaces are obtained by lifting these embeddings through the inverse map $f^{-1}$.
\par
\textbf{Energy functions}:
\textit{Map Distortion Energy}: $E_{map}(\Omega^i, \Omega^j)$ minimizes the conformal distortion between two approximate meshes $\Omega^i(V^i,T) $ and $ \Omega^j(V^j,T)$  and ensures that the map remains as conformal as possible.
A discrete conformal map is a piecewise linear map between triangulated surfaces that locally preserves angles by scaling edge lengths. Thus, preserving conformal map helps ensure that the two approximated meshes can be morphed into each other by simply contracting edges, which is the mechanism by which we actuate our robots.
For a triangle $t\in T$, $J$ is defined as the Jacobian that maps the triangle $t^i \in \Omega^i$ to $t^j \in \Omega^j$ with $A_{t^i}$ and $A_{t^j}$ as their areas. Thus, the map distortion energy is formulated as:
\begin{flalign}
    &E_{map}(\Omega^{i }, \Omega^{j}) = w_{map}\sum_{t\in T} (||J||_F  ||J^{-1}||_F - 2)(A_{t^i} + A_{t^j})\text{,}\notag
\end{flalign}
where $||J||_F  ||J^{-1}||_F$ is invariant under conformal transformation with minimum value of 2.
\par
\textit{Mesh Quality Energy}: $ E_{mesh}$ is the weighted sum of different energy terms that help in preserving the quality of approximated meshes $\Omega^i$ and is defined as:
\begin{flalign}
    E_{mesh}(\Omega^i) = w_{tri}E_{tri}(\Omega^i) + w_{bij} E_{bij}(\Omega^i) + w_{boundary}E_{bounary}(\Omega^i) +  w_{approx}E_{approx}(\Omega^i). \notag
\end{flalign}

\textit{Triangle Quality Energy}: $E_{tri}$ preserves the triangle quality of the approximating triangulations by penalizing the deviation of each triangle $ti$ from a target equilateral triangle $t^{eq}$.
We define Jacobian $J^{eqi} (t^i \rightarrow t^{eq})$ and their area as $A_{t^i}$ and $A_{t^{eq}}$.
Length of $t^{eq}$ is obtained by specifying the surface approximation parameter ($\epsilon_{approx}$) that adaptively computes the length of $t^{eq}$ based on the curvature of the mesh \cite{dunyach2013adaptive}.
$\epsilon^{eq}$ (= $0.3$) is used to tune the angle penalty to allow for variation of the standard triangle to be between equilateral and right angle triangles. This gives us the following formulation:
\begin{flalign}
    E_{tri\_area}(\Omega^{i}) &= \sum_{t\in \Omega^i} \left(\det(J^{eq})-1)\right)^2A_{t^{eq}} + \left(\det(J^{eq^{-1}})-1\right)^2A_t \text{,}\notag\\
    E_{tri\_angle}(\Omega^{i}) &= \sum_{t\in \Omega^i} max\left(0, ||J^{eq}||_F \ ||J^{eq^{-1}}||_F - (2 + \epsilon^{eq})\right)(A_t+A_{t^{eq}})\text{,} \notag\\
    E_{tri}(\Omega^{i}) &=
        E_{tri\_area}(\Omega^{i}) +
        E_{tri\_angle}(\Omega^{i})\text{.} \notag
\end{flalign}

\par
\textit{Bijective energy:} $E_{bij}$ ensures that the map between the approximated 3D mesh to its remeshed embedding remains bijective by preventing the triangles from degenerating. For a mesh $\Omega^i$, let $SA_t$ be the signed area of the triangle $t$ in its 2D embedding, the bijective energy is defined as:
\begin{flalign}
        E_{barrier}(t) &=
            \begin{cases}
                \infty,                     &\text{if} \ SA_t \leq 0\\
                -log(SA_t),    &\text{otherwise}
            \end{cases} \notag \\
        E_{bij}(\Omega^i) &= \sum_{t \in \Omega^i}E_{barrier}(t)\text{.}\notag
\end{flalign}
\newline

\textit{Boundary Energy:} $E_{boundary}$ controls the boundary of the remeshed embedding of mesh $\Omega^i$.
Since we work with disk-topology, as opposed to spherical embeddings used by \cite{schmidt_surface_2023}, all vertices of the remeshed embedding need to stay within the unit circle for the map between remeshed embedding and approximated mesh to remain bijective.
Furthermore, the remeshed embedding need to cover as much area of the unit circle to provide better approximation of the input surface. Both of these objectives are maintained by keeping the boundary vertices $\mathcal{B}^i$ of the remeshed embedding of $\Omega^i$ to lie on a unit disk. Thus giving the boundary energy as:
\[
  E_{boundary} (\Omega^{i}) = \frac{1}{\epsilon_{approx}^2.|\mathcal{B}^{ i}|}.\sum_{v \in \mathcal{B}^i} (||v||-1)^2.
\]
\par
\textit{Approximation energy:}
The vertices of $\Omega^{i}$ lie on the surface of $\Omega^{i}_0$ by construction but the reverse statement is not true.
The $E_{approx}$ measures this disparity and helps improving the approximation of $\Omega^i$ towards $\Omega^i_0$.
The vertices $v \in \Omega^i_0$ with normalized voronoi area $area(v)$ are projected on $\Omega^{i}$ from embedding space using barycentric coordinates  to obtain $\bar v$.
This gives us surface approximation energy formulation as:
\begin{flalign}
  E_{approx} (\Omega^{i}_0, \Omega^i) &= \sum_{v \in \Omega^i_0} \frac{area(v)}{\epsilon_{approx}^2}||(v-\bar{v})||. \notag
\end{flalign}
\newline
\textit{Fabrication Energy:} $E_{fab}$ consists of three energy terms designed to fulfill the fabrication constraints, i.e. bijective constrain, conformal constraint and shrinkage constraint:
\begin{flalign}
    E_{fab}(\Gamma, \Omega^i) = w_{bij}E_{bij}(\Gamma) + w_{fabmap}E_{map}(\Gamma, \Omega^i) + w_{shrink}E_{shrink}(\Gamma, \Omega^i). \notag
\end{flalign}
$E_{bij}$ and $E_{map}$ has the same formulation as before. $E_{bij}$ prevents the triangles in the 2D mesh $\Gamma$ from degenerating. $E_{map}$ optimize map from the 2D mesh $\Gamma$ to every approximated mesh $\Omega^i$ to be as conformal as possible mimicking the edge-contraction based deformation. We define the \textit{Shrinkage energy} $E_{shrink}$ to enforce the maximum shrinkage constraint (33\%) of the laser actuation setup.
For every edge $e\in T$,  the shrinkage rate is computed with the corresponding edges $e_\gamma \in \Gamma$ and $e_i \in \Omega^{ i}$ as $r(e_\gamma, e_i) = \frac{len(e_\gamma)-len(e_i)}{len(e_\gamma)}$.
$E_{sh}(r)$ is defined as a soft energy function with $\hat r = 0.33$ as the maximum shrinkage rate. $\epsilon_0$ and $\epsilon_1$ are used to soften the energy formulation. $\epsilon_2$ is initialized with $0.25$ and is gradually decreased to zero as the optimization progresses and $\epsilon_3 $ with $ 0.005$ to allow for small error buffer.
\begin{flalign}
    \epsilon_0 &=  \max(0, \ \epsilon_3 - \min_r(r)) + \epsilon_2 \text{,}\notag\\
    \epsilon_1 &= \max(0, \max_r(r) - \hat{r} + \epsilon_3) + \epsilon_2 \text{,}\notag\\
    r_0 &= \epsilon_3\text{, }r_1 =\hat r - \epsilon_3\text{,}\notag\\
    E_{sh}(r) &=
        \begin{cases}
            -log(1 + \frac{r - r_0}{\epsilon_0}) + \frac{r - r_0}{\epsilon_0}&\text{if } -\epsilon_0 + r_0 <r < r_0,\\
            0 &\text{if } r_0\leq r < r_1, \\
            -log(1 - \frac{r - r_1}{\epsilon_1}) - \frac{r - r_1}{\epsilon_1}&\text{if } r_1 \leq r < r_1+ \epsilon_1,\\
            \infty &\text{otherwise,}
        \end{cases}\notag \\
    E_{shrink}(\Gamma, \Omega^i) &= \sum_{e \in T}E_{sh}\bigg(r(\Gamma(e), \Omega^i(e))\bigg)\text{.}\notag
\end{flalign}

We weigh and normalize these energy functions and add them to obtain the combined energy function ($\mathcal{E}$) as defined in \autoref{algorithm:pseudo-code-cds}:
    \begin{flalign} \label{eq:energy-function}
        \mathcal{E} = \frac{1}{{N\choose 2}}\sum_{ij}E_{map}(\Omega^{i}, \Omega^{j})\ + \   \frac{1}{N} \sum_i E_{mesh}(\Omega^{i }) \ + \  \frac{1}{N}\sum_i E_{fab} (\Gamma, \Omega^{i }) \notag.
    \end{flalign}
\textbf{Optimization:}
We obtain the desired approximated meshes $\Omega^{i\star}$ and $\Gamma^\star$ by optimizing the above defined energy function $\mathcal{E}$ (\autoref{eq:energy-function}).
We perform the optimization in two different stages.
For each stage, the energy function $\mathcal{E}$ is optimized using the discrete continuous optimization method as described in the algorithm 1 by \cite{schmidt_surface_2023} under our fabrication constraints (\autoref{eq:fab-constraints}).
The discrete step of the algorithm applies remeshing operation including edge splits, edge collapses and edge flips to the triangulation $T$ in order to reduce the energy function.
The continuous step of the algorithm then optimizes the vertex positions of the embeddings of the approximated meshes $\Omega^i$ and the flat mesh $\Gamma$.
The weights of the first `map' stage are chosen to optimize the approximation quality of $\Omega^{i\star}$ and the weights of the second `fab' stage facilitates the optimization of $\Gamma^\star$.
Each stage is run for 100 iterations, and the tuned weights are reported in \autoref{table:optimization-parameters}.
The pseudocode in \autoref{algorithm:pseudo-code-cds} gives an overview of this process.

\begin{table*}
    \begin{center}{
    \caption{Parameters for different stages of the optimization}. \label{table:optimization-parameters}
    \resizebox{1.0\linewidth}{!}
    {
        \begin{tabular}{l||c|c|c|c|c|c|c|c|c|c|c}
        \toprule
        \rowcolor{tableheader}
        Stage &  $\epsilon_{approx}$ & $w_{map}$ &  $w_{bij}$ &$w_{boundary}$ &$w_{tri}$ & $w_{approx}$  & $w_{fabmap}$ & $w_{shrink}$ & change topology\\
        \rowcolor[HTML]{EFEFEF}
        map &  0.035 & 1.0 & $1e^{-5}$ & 1.0 & 1.0 & 1.0 & 0.0 & 0.0  & Yes\\
        \rowcolor[HTML]{EFEFEF}
        fab &    0.030 & 0.0 & $1e^{-5}$  & 1.0 & 1.0 & 1.0 & 1.0 &200.0 & No\\
        \midrule
        \end{tabular}
        }
    }
    \end{center}
\end{table*}

\paragraph{Algorithm Details for exploring the Design and Control of Terrain Navigating Robots}\label{sec:pseudo-code-dct}
Given the skeletal graph $\Gamma(V,F)$ of the robot, we aim to optimize the positions($V^\star$) and the control gait of the active edges ($C_e^\star = A^\star_e(\sin(w^\star_et + \phi^\star_e))$).
We design various reward functions to capture different target behaviors. In the walking and swimming tasks, the reward is based on the total distance the robot travels in a given direction. For the task of climbing an inclined plane, the reward is determined by the height gained by the robot.
The optimization algorithm proceeds in two steps. In the first step, we co-optimize the nodal positions and control parameters using the differential evolution algorithm \cite{storn1997differential}, implemented in \textsc{SciPy} with a population size of $5$.
This gives us an initial estimate of the vertices ($V^0$) and control gaits ($C_e^0$) of the robot.
We subsequently refine the positions and control parameters through an alternating iterative approach (five iterations). Initially, the nodal positions are optimized with a fixed control gait, followed by optimizing the control parameters while maintaining a fixed skeleton. Optimization is again performed using a differential evolution method but with a larger population size of $30$. The \frogs optimized using this method are presented in \autoref{fig:fig-6}.

\end{document}